\documentclass{article}

\usepackage{hyperref}

\usepackage[preprint]{icml2026}

\usepackage{amsmath,amsfonts,bm}

\def\eqref#1{equation~\ref{#1}}

\def\1{\bm{1}}

\def\vs{{\bm{s}}}

\DeclareMathAlphabet{\mathsfit}{\encodingdefault}{\sfdefault}{m}{sl}
\SetMathAlphabet{\mathsfit}{bold}{\encodingdefault}{\sfdefault}{bx}{n}

\usepackage{graphicx}	
\usepackage{amsmath}	
\usepackage{amssymb}	
\usepackage{amsfonts}
\usepackage{booktabs}
\usepackage{times}
\usepackage{epsfig}
\usepackage{caption}
\usepackage{float}
\usepackage{placeins}
\usepackage{color, colortbl}
\usepackage{stfloats}
\usepackage{dsfont}
\usepackage{enumitem}
\usepackage{xstring}
\usepackage{multirow}
\usepackage{xspace}
\usepackage{url}
\usepackage{subcaption}
\usepackage{xcolor}
\usepackage{lipsum} %
\usepackage[hang,flushmargin]{footmisc}
\usepackage{twemojis}
\usepackage{bbm}

\usepackage{algorithm}
\usepackage{amssymb}
\usepackage{colortbl}
\usepackage[framemethod=tikz]{mdframed}
\usepackage{tcolorbox}      %

\usepackage{xr}
\usepackage{tabularx}
\usepackage{bibunits}
\usepackage{etoolbox}

\newcommand{\myparagraph}[1]{\textbf{#1}}

\usepackage{graphicx}	
\usepackage{amsmath}	
\usepackage{amssymb}	
\usepackage{booktabs}
\usepackage{times}
\usepackage{microtype}
\usepackage{epsfig}
\usepackage{caption}
\usepackage{float}
\usepackage{placeins}
\usepackage{color, colortbl}
\usepackage{stfloats}
\usepackage{enumitem}
\usepackage{tabularx}
\usepackage{xstring}
\usepackage{multirow}
\usepackage{xspace}
\usepackage{url}
\usepackage{subcaption}
\usepackage{xcolor}
\usepackage{lipsum} %
\usepackage[hang,flushmargin]{footmisc}
\usepackage{twemojis}
\usepackage{bbm}
\usepackage[noend,ruled, vlined,algo2e]{algorithm2e}
\usepackage{algorithmic}
\usepackage{amssymb}
\usepackage{colortbl}
\usepackage[framemethod=tikz]{mdframed}
\usepackage{tcolorbox}      %
\usepackage{xr}
\usepackage{tabularx}
\usepackage{wrapfig}
\usepackage[super]{nth}
\makeatletter
\newcommand*{\addFileDependency}[1]{
  \typeout{(#1)}
  \@addtofilelist{#1}
  \IfFileExists{#1}{}{\typeout{No file #1.}}
}
\newcommand*{\newbibstartnumber}[1]{%
  \apptocmd{\thebibliography}{%
    \global\c@NAT@ctr #1\relax
    \addtocounter{NAT@ctr}{-1}%
  }{}{}%
}
\makeatother

\newenvironment{takeaway}{\begin{center}\begin{tcolorbox}[width=\linewidth, colback=cyan!9!white, colframe=black, boxrule=1pt, arc=2mm, auto outer arc, left=5pt, right=5pt, top=2pt, bottom=2pt]}{\end{tcolorbox}\end{center}}
\definecolor{iccvblue}{rgb}{0.21,0.49,0.74}

\usepackage[capitalize,noabbrev]{cleveref}

\crefname{section}{Sec.}{Secs.}
\crefname{table}{Table}{Tables}
\crefname{figure}{Fig.}{Figs.}

\newcommand{\slot}{\mathbf{S}}

\newcommand{\feat}{\mathbf{H}}

\definecolor{turquoise}{cmyk}{0.65,0,0.1,0.3}
\definecolor{purple}{rgb}{0.65,0,0.65}
\definecolor{dark_green}{rgb}{0, 0.5, 0}
\definecolor{orange}{rgb}{0.8, 0.6, 0.2}
\definecolor{dark_orange}{rgb}{0.7, 0.6, 0.3}
\definecolor{darkred}{rgb}{0.6, 0.1, 0.05}
\definecolor{blueish}{rgb}{0.0, 0.3, .6}
\definecolor{light_gray}{rgb}{0.7, 0.7, .7}
\definecolor{pink}{rgb}{1, 0, 1}
\definecolor{cyan}{rgb}{0., 1, 1}

\definecolor{myblue}{RGB}{115, 191, 249}
\definecolor{mypink}{RGB}{241, 154, 200}
\definecolor{myyellow}{RGB}{249, 219, 88}
\definecolor{myturquoise}{RGB}{152, 249, 234}
\definecolor{mysalmon}{RGB}{237, 110, 88}
\definecolor{mygreen}{RGB}{164, 247, 105}
\definecolor{myblueish}{RGB}{213, 228, 247}
\definecolor{checkyes}{rgb}{1.0, 1.0, 1.0}
\definecolor{crossno}{rgb}{1.0, 1.0, 1.0}

\newcommand{\band}{\rowcolor{myblue!25}}

\definecolor{best}{HTML}{d7191c}
\definecolor{secondbest}{HTML}{0571b0}

\newcommand{\best}[1]{\textbf{#1}}
\newcommand{\secondbest}[1]{\underline{#1}}

\newcommand{\cmark}{\text{\ding{51}}}%
\newcommand{\xmark}{\text{\ding{55}}}%

\definecolor{citecolor}{RGB}{0, 113, 188}

\Crefname{section}{Section}{Sections}
\Crefname{table}{Table}{Tables}

\crefname{figure}{Fig.}{Figs.}
\makeatletter
\DeclareRobustCommand\onedot{\futurelet\@let@token\@onedot}
\def\@onedot{\ifx\@let@token.\else.\null\fi\xspace}
\newcommand{\ourmethod}{mFRESA\xspace}
\newcommand{\llava}{LLaVA\xspace}
\newcommand{\tpone}{localization fragmentation\xspace}
\newcommand{\tptwo}{representation fragmentation\xspace}

\def\eg{\emph{e.g}\onedot} 
\def\ie{\emph{i.e}\onedot} 
\def\cf{\emph{cf}\onedot} 
\def\etc{\emph{etc}\onedot} \def\vs{\emph{vs}\onedot}

\renewcommand{\paragraph}[1]{\vspace{1.25mm}\noindent\textbf{#1}}

\newcolumntype{x}[1]{>{\centering\arraybackslash}p{#1pt}}
\newcolumntype{y}[1]{>{\raggedright\arraybackslash}p{#1pt}}
\newcolumntype{z}[1]{>{\raggedleft\arraybackslash}p{#1pt}}

\usepackage[per-mode=symbol, group-digits=integer, group-minimum-digits = 4, detect-all=true, input-symbols={()}, table-auto-round]{siunitx}

\usepackage[normalem]{ulem}
\robustify\uline

\NewDocumentCommand{\B}{}{\bfseries}
\NewExpandableDocumentCommand{\U}{m}{#1\pU{}{#1}}
\NewExpandableDocumentCommand{\UB}{m}{\B#1\pU{\B}{#1}}
\NewDocumentCommand{\pU}{mm}{%
  \llap{\uline{\phantom{#1\num{#2}}}}%
}

\newcommand{\bs}{\mathbf{s}}

\usepackage{pifont}

\usepackage[utf8]{inputenc} %
\usepackage[T1]{fontenc}    %
\usepackage{url}            %
\usepackage{booktabs}       %
\usepackage{amsfonts}       %
\usepackage{nicefrac}       %
\usepackage{microtype}      %
\usepackage{xcolor}         %

\icmltitlerunning{%
Evaluating Object-Centric Models beyond Object Discovery}
\begin{document}

\twocolumn[
  \icmltitle{%
  Evaluating Object-Centric Models beyond Object Discovery}
  \icmlsetsymbol{equal}{*}
    \begin{icmlauthorlist}
    \icmlauthor{Krishnakant Singh}{tud}
    \icmlauthor{Simone Schaub-Meyer}{tud,hess,zuse}
    \icmlauthor{Stefan Roth}{tud,hess,zuse}
  \end{icmlauthorlist}
  \icmlaffiliation{tud}{Department of Computer Science, TU Darmstadt}
  \icmlaffiliation{hess}{hessian.AI}
  \icmlaffiliation{zuse}{Zuse School ELIZA}
  \icmlcorrespondingauthor{Krishnakant Singh}{krishnakant.singh@visinf.tu-darmstadt.de}
  \icmlkeywords{Machine Learning, ICML}
  \vskip 0.3in
]

\printAffiliationsAndNotice  %

\begin{abstract}
Object-centric learning (OCL) aims to learn structured scene representations that support compositional generalization and robustness to out-of-distribution (OOD) data. However, OCL models are often not evaluated regarding these goals. Instead, most prior work focuses on evaluating OCL models solely through object discovery and simple reasoning tasks, such as probing the representation via image classification. %
We identify two limitations in existing benchmarks: \emph{(1)} They provide limited insights on the representation usefulness of OCL models,
and \emph{(2)} localization and representation usefulness are assessed using disjoint metrics. To address \emph{(1)}, we use instruction-tuned VLMs as evaluators, enabling scalable benchmarking across diverse VQA datasets to measure how well VLMs leverage OCL representations for complex reasoning tasks.
To address \emph{(2)}, we introduce a unified evaluation task and metric that jointly assess localization (\emph{where}) and representation usefulness (\emph{what}), thereby eliminating inconsistencies introduced by disjoint evaluation. Finally, we include a simple multi-feature reconstruction baseline as a reference point.
\end{abstract}

\section{Introduction}
\label{sec:intro}
Object-centric learning (OCL) aims at decomposing a scene into a set of latent object representations. 
Thereby, OCL methods aim to enable vision systems to reason about scenes by representing them as sets of constituent objects, akin to how humans reason about the world \cite{baillargeon1985object,spelke1990principles,teglas2011pure}.
Reasoning at the level of objects is thought to enable compositional or systematic generalization \cite{greff2020binding,wiedemer2024provable,kapl2025object}, improve robustness to out-of-distribution (OOD) samples \cite{dittadi2021generalization,arefin2024unsupervised}, and support causal reasoning \cite{scholkopf2021causal,mansouri2024object}.
Among various OCL approaches \cite{greff2019multi,engelcke2020genesis,lin2020space}, slot attention-based methods \cite{locatello2020object} have gained popularity for their strong performance on real-world data \cite{everingham2010pascal,lin2014microsoft}. 

\begin{figure}
    \centering
    \includegraphics[width=\linewidth]{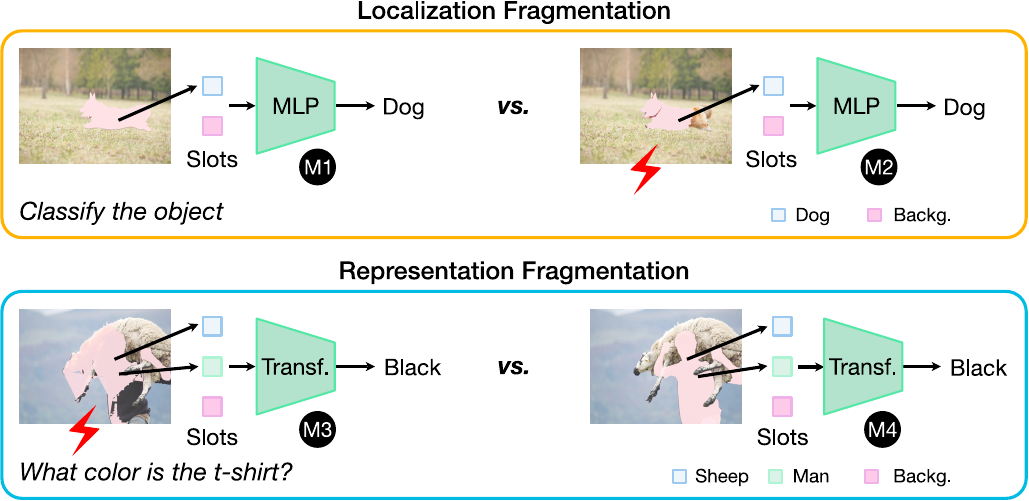}
\caption{\textbf{Issues with disjoint evaluation.}
Disjoint metrics ignore localization and representation fragmentation. \emph{(top)} Models M1 and M2 obtain the same classification score despite M1 localizing the object more accurately (localization fragmentation). \emph{(bottom)} Transformer-based probing for VQA tasks does not attribute answers to specific slots; hence, model M4 (correct answer from correct slot) is scored the same as M3, which answers correctly using the wrong slot (representation fragmentation).}
\vspace{-0.75em}
\label{fig:intro:motivation}
\end{figure}

Existing evaluation benchmarks for OCL suffer from two key limitations:
\emph{(1) Limited evaluation on complex reasoning tasks.}
Most OCL models are typically evaluated on an unsupervised object discovery task. This serves as a poor proxy \cite{rubinstein2025we} for evaluating the broader goals behind OCL models, such as OOD generalization, counterfactual reasoning, and compositional generalization to novel scenes. Using the common linear probing for each such capability is cumbersome and does not scale, as it requires repeated retraining to evaluate each capability. Moreover, some qualities (\eg, counterfactual reasoning) are not naturally expressed as small closed-set classification problems, making linear probing ill-suited and potentially ambiguous. Other proposed solutions for evaluating OCL performance on visual question-answering tasks \cite{mamaghan2024exploring} fall short, as they also require retraining to assess different reasoning abilities of OCL models. 
\emph{(2) Disjoint evaluation metrics.}
Localization and representation usefulness are typically evaluated using separate metrics. This disjoint evaluation can lead to inconsistencies such as \emph{localization fragmentation}, where a model may capture semantics well but fail to localize the object correctly, and \emph{representation fragmentation}, where a single object is encoded by multiple slots (see \cref{fig:intro:motivation}).

To address \emph{Limitation 1}, we propose a scalable evaluation framework for OCL that leverages visual instruction tuning to convert a large language model (LLM) into a vision-language model (VLM) and uses an object-centric model as the vision encoder. This enables a \emph{zero-shot evaluation} using diverse visual question-answering (VQA) benchmarks, designed to test broad visual reasoning capabilities without task-specific training for each new benchmark. Importantly, this evaluation measures how well a VLM leverages object-centric representations across visual reasoning tasks, serving as a practical proxy for the representations' utility. 
Using VQA-based evaluation alone is not yet sufficient, as it does not account for the issue of disjoint assessment of localization and representation usefulness. To address \emph{Limitation 2}, we introduce a unified evaluation task using our enhanced version of the GQA dataset \cite{hudson2019gqa} and a novel attribution-aware grounded accuracy (AwGA) metric that jointly assesses the object localization and the usefulness of OCL models' representations.

To summarize, our contributions are:
\emph{(i)} We benchmark multiple OCL methods by evaluating their usefulness in a VLM across diverse VQA benchmarks designed to test broad visual reasoning abilities; we do so in a zero-shot fashion without requiring retraining.
\emph{(ii)} We introduce attribution-aware grounded accuracy (AwGA), a metric that jointly evaluates the ``what'' and ``where'' properties of OCL models, addressing localization and representation fragmentation.
\emph{(iii)} We show that our VLM-based evaluation and AwGA yield consistent model rankings across different LLM backbones and connectors.
\emph{(iv)} We show that multi-feature reconstruction consistently improves the usefulness of object-centric representations under our VLM-based evaluation.

\section{Related work}
\textbf{Object-centric learning (OCL)} shares certain goals with other object representation learning approaches, such as CLIP \cite{radford2021learning}, DINO \cite{caron2020unsupervised,oquab2023dinov2}, and VQ-VAE \cite{van2017neural}. However, unlike them, OCL methods aim to learn \emph{object-level} latent representations %
that enable robust and compositional scene understanding \cite{dittadi2021generalization,wiedemer2024provable}. Early OCL models relied on VAE-based architectures \cite{burgess2019monet,greff2019multi} and suffered from scalability issues, which slot attention \cite{locatello2020object} addressed via iterative attention-based clustering. \citet{seitzer2022bridging} extended slot attention to real images by reconstructing DINO features \cite{caron2020unsupervised}. Modern OCL models can be grouped by reconstruction target: image-based approaches \cite{jiang2023object,wu2023slotdiffusion,singh2024guided,akan2025slot} reconstruct pixels using strong decoders such as StableDiffusion \cite{rombach2022high}, while feature-based models \cite{seitzer2022bridging,kim2024bootstrapping,kakogeorgiou2024spot} reconstruct self-supervised encoder features. 
\begin{table}[t]
\centering
\scriptsize
\caption{\textbf{Issues with existing evaluation protocols.} Linear and transformer-based probes require retraining for each property, leading to high amortized cost (cost per evaluation), and linear probes are not applicable to tasks requiring open-ended outputs (Open eval.). VLM probes enable diverse evaluations in a zero-shot manner but still suffer from issues of disjoint evaluation (Rep. and Loc. fragmentation). Combining AwGA with VLMs provides a unified evaluation that penalizes OCL methods for rep. and loc. fragmentations.}
\label{tab:related_work:comparison}
\begin{tabularx}{\linewidth}{@{}l*{4}{X}@{}}
\toprule
Evaluation Protocol & Amort. cost & Open eval. & Rep. frag. & Loc. frag. \\
\midrule
    Linear probes              & High & \textcolor{red}{\xmark}  & \textcolor{red}{\xmark} & \textcolor{red}{\xmark} \\
    Transformer probes  & High & \cmark & \textcolor{red}{\xmark} & \textcolor{red}{\xmark} \\
    VLM probes \emph{(ours)}          & Low  & \cmark & \textcolor{red}{\xmark} & \textcolor{red}{\xmark} \\
    VLM probes + AwGA \emph{(ours)}   & Low & \cmark & \cmark & \cmark \\
\bottomrule
\end{tabularx}
\vspace{-0.75em}
\end{table}

\myparagraph{Evaluation of OCL models.}
Unsupervised object discovery (UOD) is the most common evaluation for OCL. However, as \citet{rubinstein2025we} argue, UOD is a poor proxy for evaluating OCL models since it does not assess key goals such as compositional generation, counterfactual reasoning, and OOD generalization. Beyond UOD, prior work often evaluates OCL representations via linear probing on downstream property prediction tasks \cite{locatello2020object,jiang2023object,singh2024guided}, but this setup is ill-suited for complex, varied reasoning tasks: \emph{(i)} it incurs high amortized cost, as each new benchmarking task requires retraining the probes, and \emph{(ii)} many desirable qualities (\eg, counterfactual reasoning) are difficult to express as closed-set classification tasks. To address this, \citet{mamaghan2024exploring} proposed using a VQA-based evaluation with transformer probes. However, their approach still requires expensive retraining to benchmark each new capability \citep[\eg, compositional generalization;][]{kapl2025object}, resulting in a high amortized cost. Moreover, repeated training for different benchmarks requires a cumbersome hyperparameter search (see \cref{tab:related_work:comparison}).

Inspired by \citet{tong2024cambrian}, we instead use instruction-tuned vision-language models (VLMs) as probes, enabling \emph{scalable zero-shot evaluation}. Unlike \citet{mamaghan2024exploring}, our work does not require training for each new property evaluation.
In particular, our protocol measures how effectively object-centric representations can be aligned with and exploited by a language model across diverse tasks, providing a practical measure of their downstream utility in multimodal systems. However, VQA-based protocols typically evaluate representation and localization separately, overlooking fragmentation effects. We address this by proposing a new evaluation protocol that uses a novel attribution-aware grounded accuracy (AwGA) metric to jointly evaluate localization and representation (see \cref{fig:intro:motivation} and \cref{tab:related_work:comparison}).

\section{Benchmarking beyond object discovery}
\subsection{Preliminaries}
\textbf{Slot attention} \citep[\textbf{SA};][]{locatello2020object} is an iterative refinement framework that decomposes an image into a set of object-centric slots. Given an encoder feature map $\feat$, the slot-attention module groups it into $k$ slot vectors $\slot=\{\bs_1,\dots,\bs_k\}$. At each iteration $t$, slots $\slot^{t}$ are updated via dot-product attention \cite{vaswani2017attention} between the previous slots $\slot^{t-1}$ and features $\feat$, where the softmax is taken over slots (instead of keys), inducing competition and binding of slots to objects. Typically, SA-based methods use $k=7$ slots for real-world scenes \cite{lin2014microsoft}, as this setting performs well on common downstream tasks.

\begin{figure*}
    \centering
    \includegraphics[width=0.85\textwidth]{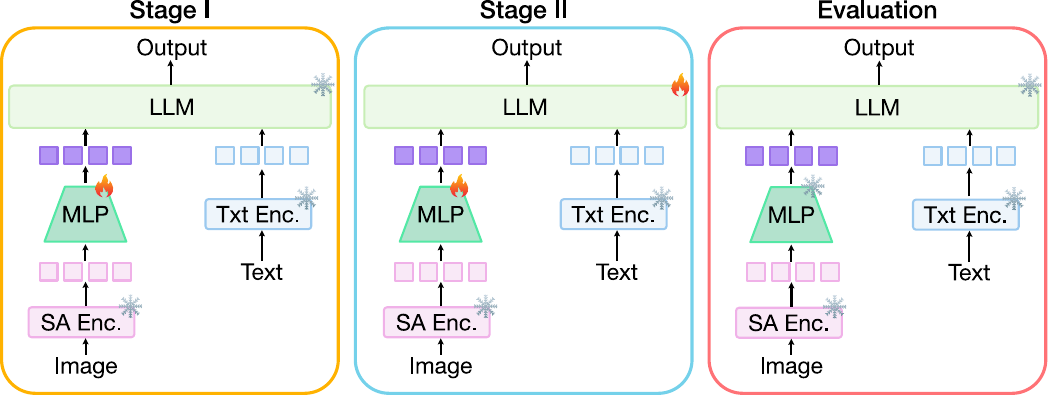}
    \caption{\textbf{Training and evaluation setup.} Our training is akin to \llava \cite{liu2023visual}. In \emph{Stage I}, only the MLP connector is trained on the pre-training dataset. This aligns the slot embeddings with the language model's embedding space. In \emph{Stage II}, the MLP network and the language model are trained on the instruction-tuning dataset from \llava. This enables the language model to follow instructions and perform tasks based on slots as visual tokens. \emph{Evaluation} is performed in a zero-shot fashion on various VQA benchmarks, where the text is encoded via a text encoder, and images are encoded using the slot-attention model. The evaluation tests how well the connector and LLM networks utilize the slot embeddings for answering the provided questions.}
    \label{fig:method:llava_training}
    \vspace{-0.75em}
\end{figure*}

\subsection{Using VLMs as evaluators} \label{sec:method:vlm}
One of the core contributions of our work is to benchmark OCL models by measuring their utility in a VLM across diverse VQA benchmarks designed to test broad visual reasoning capabilities. Extending existing protocols, such as linear probing and transformer-based probing \cite{mamaghan2024exploring}, requires training probes from scratch for each benchmark (see \cref{tab:related_work:comparison}). To alleviate this issue, we take inspiration from \citet{tong2024cambrian} and use instruction-tuned VLMs as evaluators, replacing the standard vision encoder (\eg, CLIP \cite{radford2021learning} or DINOv2 \cite{oquab2023dinov2}) with the OCL encoders.

Our VLM-based evaluation scheme can be written as a function composition $f(g(\slot))$, where $f$ denotes an LLM and $g$ denotes a connector network (\eg, 2-layer MLP) that connects the LLM to the slot representation $\slot$. Previous evaluation protocols such as linear probing ($f=\mathbf{I}$, $g=1\text{-layer MLP}$) and transformer-based probes \citep[$f=\mathbf{I}$, $g=n\text{-layer transformer}$;][]{mamaghan2024exploring}, where $\mathbf{I}$ denotes the identity function, are special cases of our generalized evaluation protocol. Since both $g$ and $f$ are trained, our benchmark measures the \emph{usefulness} of OCL representations on diverse VQA benchmarks in a zero-shot fashion without requiring retraining for each task. The performance of the OCL model is influenced not only by the visual information in the object-centric representation but also by the ease with which a VLM can align and exploit it to answer textual questions. 
Note that training linear probes or transformer-based probes similarly evaluate not only what information is present in the representation, but also how easily it can be extracted by the chosen probe class. Our VLM-based protocol follows the same principle, but uses a stronger probe (an instruction-tuned VLM), enabling scalable evaluation across diverse VQA tasks.
We follow the architecture and training protocol of \llava{} \cite{liu2023visual} for learning a vision-language model, using object-centric models as vision encoders. 

The training process (see \cref{fig:method:llava_training}) has two stages: 
\emph{(i)} In \emph{Stage I}, the slot embeddings are projected by a connector network to align with the space of text embeddings. The text (questions) is tokenized and embedded using the LLM's embedder module. Only the connector network is trained for one epoch on the \llava\ 558K pre-training dataset in this stage. 
\emph{(ii)} In \emph{Stage II}, both the LLM and connector networks are trained with the \llava\ 665K instruction tuning dataset, which comprises multimodal samples created via GPT-4's responses \cite{achiam2023gpt} to images. For more details, see \cite{liu2023visual}.
The instruction-tuning phase helps the model follow instructions more reliably and improves the VLM's ability to follow instructions and leverage the slot encodings to accurately respond to user prompts.

\subsection{Joint evaluation protocol -- Unifying `what' and `where'}\label{sec:exp:new_metric} \label{sec:method:awga}

With VLM probes, we can evaluate the usefulness of many object-centric models on a diverse set of VQA-based benchmarks \emph{in a zero-shot manner at evaluation time}, without training task-specific probes for each benchmark. This allows us to assess how well slot representations support a broad range of capabilities for which OCL models were originally proposed. However, relying solely on VQA-based evaluation introduces both localization and representation fragmentation (see \cref{fig:intro:motivation}). Thus, a need exists for a novel evaluation protocol that jointly evaluates and penalizes localization and representation fragmentations. This requires access to a dataset with grounding masks, composed of the masks of all objects required to answer a question.

\begin{figure}
\centering
\includegraphics[width=\linewidth]{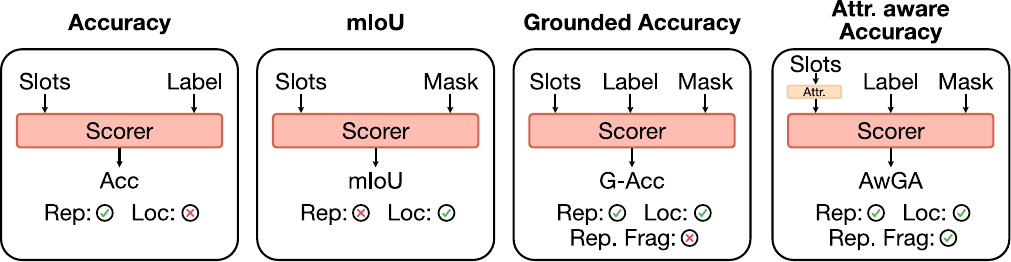}
\caption{\textbf{Metrics for evaluating OCL models.} Accuracy and mIoU evaluate representation usefulness and localization separately, while grounded accuracy allows for a joint evaluation, addressing localization fragmentation but overlooks representation fragmentation. Our proposed AwGA jointly evaluates both and penalizes both fragmentation types.}
\vspace{-0.75em}
\label{fig:exp:metrics_visual}
\end{figure}

A way to account for \tpone when evaluating different models with VQA tasks is to use the grounded accuracy \citep[G-Acc;][]{hudson2019gqa}, which is defined as 
\begin{equation}
    \text{G-Acc} = \frac{1}{N} \sum_{i=1}^N \mathbbm{1}(\hat{y}=y) \; \text{mIoU} (\mathcal{A}_{\text{pred}}, \mathcal{G}_{\text{GT}}).
\end{equation}
Here, $\mathcal{A}_\text{pred}$ and $\mathcal{G}_{\text{GT}}$ denote the mask predicted from the slots and the ground-truth (GT) grounding masks. $y$ and $\hat{y}$ denote the GT and predicted answer; $\mathbbm{1}$ is the indicator function.
G-Acc correctly penalizes \tpone; however, it does not consider which slot was used to answer the question. Specifically, G-Acc does not penalize \tptwo, \ie, when a model distributes an object's representation across multiple slots (see \cref{fig:exp:metrics_visual}).

To resolve this and enable the joint evaluation of localization and representation usefulness, we propose \emph{AwGA}, an attribution-aware grounded accuracy metric that penalizes a model for committing both localization and representation fragmentation (\cref{fig:exp:metrics_visual}). 
Our AwGA metric first computes an attribution score for each slot with respect to the predicted answer \cite{simonyan2013deep}. We then select the $K$ slots with the highest attributions and compute the mean intersection over union (mIoU) using the \emph{union} of their predicted masks. For each question, $K$ is set to the number of objects in the grounding mask and is \emph{not} a hyperparameter. This way, the overlap is computed only for the slots most responsible for answering the question. AwGA is formally written as 
\begin{equation}
    \text{AwGA}= \frac{1}{N} \sum_{i=1}^N \mathbbm{1}(\hat{y}=y) \;
    \text{mIoU} (\text{TopK}(\mathcal{A}_{\text{pred}}), \mathcal{G}_{\text{GT}}).    
\end{equation} 

For computing each attribution, we simply use the gradient of each slot with respect to the loss function \cite{simonyan2013deep,springenberg2014striving}. In particular, we compute the sensitivity ($\frac{\partial y}{\partial \bs_i}$) of the output $y=\mathit{f}(\mathit{g}(\slot))$ with respect to each slot $\mathbf{s}_i$.

\section{Analysis of OCL methods}
\paragraph{Utility \vs grounded faithfulness.}
We first benchmark the \emph{utility} of OCL models in a VLM across diverse VQA benchmarks designed for evaluating broad visual reasoning capabilities (\cref{tab:exp:main_cmp_vqa,tab:exp:robust_vqa}). 
While these evaluations quantify the usefulness of OCL models across diverse complex reasoning tasks, they do not address issues arising from disjoint evaluation (fragmentation failure modes). 
Additionally, \cref{sec:exp:correlation} shows that using object discovery as a measure is a poor proxy for an OCL method's downstream capabilities, underscoring the need for a unified metric.
We therefore propose a joint evaluation protocol based on our enhanced GQA dataset and using our AwGA metric, simplifying the evaluation of \emph{what} and \emph{where} with a single score (\cref{sec:exp:awga}).

\myparagraph{Baselines.} The original goal of object-centric learning (OCL) has been to obtain object-centric representations in an \emph{unsupervised} manner; we thus focus our evaluation on unsupervised OCL methods. We take state-of-the-art baselines for real-world datasets \citep[\eg, COCO;][]{lin2014microsoft}, including SPOT \citep{kakogeorgiou2024spot}, Slot\-Diffusion \citep{wu2023slotdiffusion}, StableLSD \citep{jiang2023object}, FT-DINOSAUR \citep{didolkar2025zero}, and DINOSAUR \citep{seitzer2022bridging}. Whenever available, we use the authors’ released checkpoints; StableLSD and DINOSAUR are retrained from scratch using their official scripts. We also include a DINOSAURv2 baseline, which replaces the original DINO \citep{caron2021emerging} backbone with DINOv2 \citep{oquab2023dinov2} and a vector-quantized version of DINO (VQDINO$_{\text{MLP}}$) introduced by \citet{zhao2025vvo}.

\myparagraph{Improved baseline.} 
Existing OCL models typically use either feature reconstruction \citep{seitzer2022bridging,kakogeorgiou2024spot} or image reconstruction \citep{jiang2023object,wu2023slotdiffusion} as their training target. For completeness, we include a simple improved baseline, \ourmethod, which combines multiple reconstruction targets: image pixels, DINOv2 features, and HOG features \citep{dalal2005histograms}. Concretely, \ourmethod builds on StableLSD \citep{jiang2023object} and adds two lightweight decoders: \emph{(1)} a feature decoder that reconstructs DINOv2 features, akin to \citet{seitzer2022bridging}, and \emph{(2)} a three-layer MLP that reconstructs patch-level HOG features \citep{dalal2005histograms}. For further details, see \cref{sec:appx:method_train_detail}. While not a main contribution, this baseline provides a useful reference and demonstrates that combining complementary reconstruction signals can yield improved downstream utility and AwGA scores. 

\myparagraph{Training details for VLM-based evaluation of OCL methods.} \label{sec:exp:training_details}
We use Phi2-3B \citep{javaheripi2023phi} and Qwen2-7B \citep{an2024qwen2} as language models in our VLM evaluation setup. We use a 2-layer MLP with GeLU activations \citep{hendrycks2016gaussian} as the connector network. Pre-training is performed with a batch size of \num{256} and a learning rate of \num{1e-3}, followed by fine-tuning with a batch size of \num{128} and a learning rate of \num{2e-5}, using AdamW \citep{loshchilov2017fixing} throughout. The maximum sequence length is set to 2048 tokens. During evaluation, we follow \llava and use greedy decoding (temperature 0, beams 1). All training and architectural settings are held fixed across LLMs and vision encoders, ensuring that the only variability comes from the choice of vision encoder (\ie, slot attention) module. VLM training is performed on 8×A100 GPUs (80 GB each). Training time varies across OCL encoders, with the longest runs occurring with mFRESA and StableLSD. For these models, pre-training requires approximately 4 hours, and fine-tuning takes roughly 20–24 hours.

\subsection{Standard perception evaluation} \label{sec:exp:eval_standard}
\newcolumntype{K}{>{\centering\arraybackslash}X}
\begin{table*}[tb]
    \centering
    \scriptsize
    \setlength{\tabcolsep}{1pt}
  \caption{\textbf{VQA comparison of OCL and foundational models.} Blind-VLM replaces vision tokens with random noise and serves as a lower bound, isolating the contribution of the vision encoder. DINOv2 provides an upper-bound reference for self-supervised representations, compared against slot-attention models using feature (\nth{2} group) or image (\nth{3} group) reconstruction. We highlight the \textbf{best} and \secondbest{second-best} SA models and report VQA accuracy (\%, $\uparrow$); for MME, only perception tasks are included ($\uparrow$, max 2000). Despite using only 7 vision tokens, OCL models remain competitive with DINOv2, which uses 196 tokens. \ourmethod{} a hybrid feature and image-reconstruction method outperforms other OCL methods on most benchmarks.}
  \vspace{-0.5em}
    \label{tab:exp:main_cmp_vqa}
    \begin{tabularx}{\textwidth}{@{}l*{11}K@{}}
    \toprule
    \textbf{LLM} & \multicolumn{5}{c}{\textbf{Phi2}} & \multicolumn{5}{c}{\textbf{Qwen2-7B}} \\
    \cmidrule(lr){2-6}\cmidrule(lr){7-11}
    \textbf{Dataset} & \textbf{GQA} & \textbf{POPE} & \textbf{MME} & \textbf{MMVET} & \textbf{VQAv2} & \textbf{GQA} & \textbf{POPE} & \textbf{MME} & \textbf{MMVET} & \textbf{VQAv2} \\
    \midrule
    Blind-VLM & 38.20 &  64.92 & 667.38 & 13.2  & 44.71 &  40.32 & 65.12 & 686.29 &  15.0 & 45.76\\
    DINOv2 & 57.77 & 82.01 & 1279.21 & 22.6 & 71.15 & 61.83 & 83.31 & 1388.19 & 23.6 & 74.86 \\
    \midrule
    DINOSAUR & 49.71 & 78.72 & 1047.34 & 17.5 & 58.41 & 53.63 & 79.54 & 1178.87 & 15.9 & 61.98 \\
    DINOSAURv2 & \secondbest{53.23} & \secondbest{81.76} & 1122.73 & \best{18.9} & \secondbest{63.84} & \secondbest{56.32} & \secondbest{82.49} & 1224.84 & 17.7 & 66.23 \\
    VQDINO$_{\text{MLP}}$ & 52.69 & 81.48 & \secondbest{1167.08} & \secondbest{18.8} & 63.22 & 
    56.2 & 82.36 & 1229.43 & 18.8 & \secondbest{67.28}  \\
    FT-DINOSAUR  & 52.22 & 81.45 & 1004.94 & 15.1 & 60.97 & 56.15 & 81.85 & \secondbest{1242.25} & 17.2 & 66.09 \\ 
    SPOT & 51.06 & 79.74 & 1069.80 & 17.4 & 60.94 & 54.94 & 80.24 & 1169.11 & 17.8 & 65.37 \\ 
    \midrule
    Slot Diffusion & 50.00 & 79.77 & 1090.10 & 18.5 & 59.65 & 53.93 & 79.91 & 1171.83 & \secondbest{18.9} & 63.47 \\ 
    StableLSD & 51.45 & 81.51 & 
    1129.08 & 17.8 & 62.06 & 55.96 & 81.54 & 1239.48 & 18.6 & 66.67 \\
    \midrule
    \band \ourmethod \emph{(ours)} & \best{53.90} & \best{82.12} & \best{1187.05} & 18.5 & \best{65.58} & \best{58.28} & \best{82.74} & \best{1283.48} & \best{19.3} & \best{69.93} \\
    \bottomrule
    \end{tabularx}
    
\end{table*}

Using instruction-tuned VLMs as evaluators enables scalable benchmarking of vision encoders across a wide range of tasks represented in a VQA setting. Since our goal is to assess the representational utility of object-centric vision tokens, we focus on image-centric perception benchmarks, including GQA \citep{hudson2019gqa}, VQAv2 \citep{goyal2017making}, MME \citep{fu2024mme}, and MM-Vet \citep{yu2023mm}. To evaluate whether object-centric encodings can mitigate object hallucination, we also report results on POPE \citep{li2023evaluating}, which probes object presence via Boolean questions. For MME, we report only perception tasks, as these are most relevant to our setting. To quantify the extent to which performance depends on visual tokens (rather than language priors), we also include a \emph{Blind-VLM} baseline, where vision tokens are replaced with random noise.

As shown in \cref{tab:exp:main_cmp_vqa}, despite using far fewer visual tokens (7 \vs~196 for DINOv2), OCL models perform competitively with DINOv2, a strong self-supervised vision encoder. Interestingly, FT-DINOSAUR — the leading OCL model for object discovery — underperforms the older DINOSAURv2 on nearly all benchmarks, suggesting that object discovery scores are a poor proxy for the utility of OCL representations \citep[\cf][]{rubinstein2025we} also in our VLM-based visual reasoning benchmarks. 

Combining reconstruction targets (\ourmethod) improves performance across most benchmarks (except MM-Vet), suggesting that multi-target reconstruction can be a useful design choice for OCL.
\begin{takeaway}
    \emph{Takeaway 1.} Under our VLM-based evaluation, OCL models can be competitive with DINOv2 despite using far fewer tokens. Overall, feature-reconstruction models outperform image-reconstruction models, and a hybrid baseline (\ourmethod) yields further gains.
\end{takeaway}

\subsection{Robust perception evaluation} 
\label{sec:exp:eval_robust}
Despite competitive results, OCL models still lag in absolute performance on general perception benchmarks. We next ask whether they offer advantages on tasks where object-centric learning is conjectured to help, such as OOD generalization, compositionality, and counterfactual reasoning \citep{greff2020binding,wiedemer2024provable,kapl2025object}. 
Using VLM probes enables evaluating OCL encoders \emph{without retraining} across diverse VQA benchmarks designed to test these abilities. Results are shown in \cref{tab:exp:robust_vqa}.

\emph{Positives.}
On OOD-CV \citep{tu2024many}, which contains images with unusual textures and backgrounds, most OCL models are competitive with DINOv2 despite using far fewer visual tokens, suggesting that OCL models are robust to distribution shifts. 
We also find that on counterfactual question answering, particularly direct numeric queries (\eg, ``How many $X$ would there be if two $X$ were added/removed?''), several OCL models are competitive with or outperform DINOv2. Interestingly, for Boolean counterfactual questions, the non-visual baseline (\emph{Blind-VLM}) performs best. This suggests that many Boolean CVQA questions are structurally simple (\eg, ``Would $X$ still be true if $Y$ changed?'') and can often be answered using linguistic priors, while adding visual tokens may introduce signals that are not helpful in such cases. It should be noted that success on counterfactual QA does not necessarily imply causal understanding. Also, it can be seen that using a multi-feature reconstruction target (mostly) improves the performance on these robust perception-reasoning tasks.

\emph{Negatives.}
For compositional reasoning, we evaluate SugarCrepe \citep{hsieh2023sugarcrepe}, where the model must choose the correct caption between a true caption and a hard-negative caption generated by an LLM \citep[attribute swaps, object additions, or replacements; ][]{achiam2023gpt}. OCL models remain behind DINOv2.
We also evaluate robustness to natural adversarial examples using NaturalBench \citep{li2024naturalbench}, which contains pairs of questions and images designed so that a blind model fails (\ie, the answer changes with the image). Solving NaturalBench requires object recognition, attribute binding, and relation understanding. We again observe a substantial gap between OCL models and DINOv2, indicating that current OCL representations are less useful for this task in our VLM-based evaluation setting. Additionally, we find that feature-reconstruction-based OCL models outperform image-reconstruction-based models under our VLM probing protocol.

\begin{table*}[t]
    \centering
    \scriptsize
    \caption{\textbf{Robustness of OCL methods.} Evaluation on tasks beyond object discovery, such as OOD generalization, compositional understanding, counterfactual reasoning, \etc (accuracy in \%, $\uparrow$).  The Blind-VLM serves as a lower bound while DINOv2 serves as a reference upper bound for performance on these tasks. We highlight the \textbf{best} and \secondbest{second best} model among SA methods. The datasets evaluate the following properties: CVQA \citep{zhang2024if} -- counterfactual reasoning, OOD-CV \citep{tu2024many} -- OOD generalization, NaturalBench \citep{li2024naturalbench} -- robustness to natural adversarial examples, SugarCrepe \citep{hsieh2023sugarcrepe} -- vision-language compositionality.}
    \setlength{\tabcolsep}{2pt} %
    \label{tab:exp:robust_vqa}
    \vspace{-0.5em}
    \renewcommand{\arraystretch}{1.15} %
    \begin{tabularx}{\linewidth}{@{}l
    *{10}K
    @{}}
    \toprule
    \textbf{LLM} & \multicolumn{5}{c}{\textbf{Phi2}} & \multicolumn{5}{c}{\textbf{Qwen2-7B}} \\
    \cmidrule(lr){2-6}\cmidrule(lr){7-11}
    \textbf{Dataset} & \multicolumn{2}{c}{\textbf{CVQA}} & \textbf{OODCV} & \textbf{N. Bench} & \textbf{SugarC.} & \multicolumn{2}{c}{\textbf{CVQA}} & \textbf{OODCV} & \textbf{N. Bench} &
    \textbf{SugarC.} \\
    \cmidrule{2-3}
    \cmidrule{7-8}
    & Direct & Boolean&  &  &  & Direct & Boolean &  &  &  \\ 
    \midrule
    Blind-VLM & 28.00 & 71.41 & 50.98 & 0.42  & 49.42 & 32.69 & 65.57 & 51.01 & 0.52 & 53.29 \\
    DINOv2 & 36.96 & 63.72 & 58.00 & 8.42 & 82.05 & 45.74 & 53.54 & 58.36 & 9.89 & 88.06 \\ 
    \midrule
    DINOSAUR & 35.74 & 69.29 & 51.97 & 1.89 & 67.85 & 41.13 & \secondbest{63.72} & 52.52 & 3.95 & 72.45 \\ 
    DINOSAURv2 & 34.52 & 65.75 & 53.90 & 3.37 & \secondbest{75.98} & \secondbest{42.09} & \best{64.07} & \secondbest{56.66} & \secondbest{6.16} & 78.18 \\ 
    VQDINO$_\text{MLP}$ & 35.13 & 66.37 & 52.03 & \secondbest{3.89} & 73.05 & \best{42.34} & 55.66 & 53.18 & 6.07 & 80.16 \\
    FT-DINOSAUR & \best{39.13} & 68.85 & \secondbest{55.18} & 2.89 & 70.94 & 42.00 & 57.17 & 53.28 & 5.42 & \secondbest{81.24} \\ 
    SPOT & 36.35 & \secondbest{69.47} & 53.34 & 2.42 & 71.65 & 41.83 & 57.61 & 54.07 & 3.68 & 74.08 \\ 
    \midrule
    Slot Diffusion & 33.83 & 68.23 & 51.34 & 2.21 & 70.39 & 39.39 & 59.56 & 52.56 & 3.74 & 74.53 \\ 
    StableLSD & \secondbest{38.26} & \best{70.44} & 52.89 & 3.00 & 72.92 & 41.04 & 62.39 & 55.08 & 5.21 & 78.98 \\ 
    \midrule
    \band mFRESA \emph{(ours)} & 38.09 & 66.64 & \best{55.57} & \best{4.21} & \best{77.27} & 41.39 & 60.44 & \best{57.31} & \best{6.84} & \best{83.17} \\ 
    \bottomrule
    \end{tabularx}
\end{table*}

\begin{takeaway}
    \emph{Takeaway 2.} Under our VLM-probing evaluation, OCL models are competitive on OOD generalization and numeric counterfactual reasoning, but lag behind DINOv2 on compositional and natural adversarial benchmarks. Feature-reconstruction OCL generally outperforms image-reconstruction on robustness tasks.
\end{takeaway}

\subsection{Are object discovery metrics predictive of downstream utility?} \label{sec:exp:correlation}
Unsupervised object discovery (UOD) metrics such as mean best overlap \citep[mBO;][]{pont2016multiscale} and mean intersection over union (mIoU) are widely used to evaluate slot-attention methods. Yet, it remains unclear whether higher UOD scores imply greater utility of object representations for diverse reasoning tasks \citep[\cf][]{rubinstein2025we}. In \cref{tab:exp:unreliablity_od}, we compare several OCL methods across UOD metrics, general VQA performance, adversarial robustness, and compositional reasoning. 
We find that UOD metrics (mIoU and mBO) correlate poorly with the usefulness of slot representations under our VLM-probing evaluation. For instance, FT-DINOSAUR, a leading OCL model for object discovery, performs worse than DINOSAURv2 on general VQA tasks and robustness benchmarks (compositional reasoning and natural adversarial robustness). One plausible explanation is that FT-DINOSAUR finetunes the DINOv2 encoder, whereas other models keep it frozen; finetuning on comparatively small datasets such as COCO can reduce generalization \citep{mukhoti2024finetuning}, which may negatively affect the learned object representations required for reasoning tasks.
\begin{table}[t]
\centering
\scriptsize
\setlength{\tabcolsep}{2pt}
\centering
\caption{\textbf{Object discovery (OD) and representational quality are uncorrelated.} FT-DINOSAUR scores highest on OD metrics (mBO$_i$, mIoU) but underperforms on various VQA tasks (all in \%,~$\uparrow$). All methods use DINOv2 as the backbone. The Spearman's rank correlation between accuracy (VQAv2) and mIoU for these models is $-0.2$, indicating a negative correlation.} 
\label{tab:exp:unreliablity_od}
\vspace{-0.5em}
\begin{tabularx}{\linewidth}{@{}l*{5}K@{}}
\toprule
\textbf{Dataset} & {\textbf{VQAv2}} & {\textbf{Nat. Bench}} & {\textbf{Sugar C.}} & \multicolumn{2}{c}{\textbf{COCO}}  \\\cmidrule(lr){2-4}\cmidrule(lr){5-6}
\textbf{Metric} & \multicolumn{3}{c}{\textbf{accuracy}} & {\textbf{mIoU}} & {\textbf{mBO$_i$}} \\
\midrule
DINOSAURv2 & \U{63.84} & \U{3.37} & \U{75.98} & 27.25 & 28.42 \\
FT-DINOSAUR & 60.97 & 2.89 & 70.94 & \B 34.52 & \B 36.08 \\
StableLSD & 62.06 & 3.00 & 72.92 & 24.52 & 25.72 \\
\midrule
\band \ourmethod \emph{(ours)} & \B 65.75  & \B 4.11  & \B 77.17 & \U{30.60} & \U{32.17} \\
\bottomrule
\end{tabularx}
\end{table}

\begin{takeaway}
    \emph{Takeaway 3.} In our VLM-based evaluation setting, object discovery metrics are weak proxies for the usefulness of object-centric representations, motivating metrics that jointly evaluate localization \emph{and} usefulness of OCL representations.
\end{takeaway}

\begin{table}[t]
    \centering
    \scriptsize
    \caption{\textbf{Performance comparison of different models using G-Acc and AwGA metrics} (all in \%, $\uparrow$). As seen, the mIoU and Acc. measures exhibit a weak Spearman correlation (Phi2: 0.35 and Qwen2: 0.50), indicating that either metric alone is a poor proxy for evaluating both the localization and the usefulness of the representation of OCL methods. G-Acc and AwGA jointly evaluate both; however, AwGA additionally penalizes representation fragmentation (also see \cref{fig:exp:qual_samples}).}
    \vspace{-0.5em}
    \label{tab:exp:awga_metric}
    \setlength{\tabcolsep}{2pt} %
    \begin{tabularx}{\linewidth}{@{}l
    *{8}K
    @{}}
    \toprule
    \textbf{LLM} & \multicolumn{4}{c}{\textbf{Phi2}} & \multicolumn{3}{c}{\textbf{Qwen2-7B}} \\
    \cmidrule(lr){2-5}\cmidrule(lr){6-8}
        \textbf{Metric} & \textbf{mIoU} & \textbf{Acc.} & \textbf{G-Acc.} & \textbf{AwGA} & \textbf{Acc.} & \textbf{G-Acc.} & \textbf{AwGA}\\ 
        \midrule
        DINOSAUR & 50.52 & 60.13 & 30.80 & 11.64 & 64.05 & 32.71 & 13.47 \\
        DINOSAURv2 & 47.99 & \secondbest{66.27} & 32.40 & 11.92 & 68.54 & 33.18 & 12.44 \\ 
        VQDINO$_\text{MLP}$ & 48.50 & 65.55 & 32.27 & 12.13 & 68.18 & 34.13 & 13.47 \\
        FT-DINOSAUR & \best{59.09} & 61.05 & 33.94 & 13.25 & 65.21 & 36.28 & \secondbest{15.36} \\ 
        SPOT & 53.76 & 64.08 & \secondbest{38.45} & \secondbest{12.81} & 68.32 & \best{41.45} & 15.14 \\ 
        \midrule
        Slot Diffusion & 54.91 & 61.54 & 34.39 & 12.42 & 65.75 & 36.91 & 13.85 \\ 
        StableLSD & 47.94 & 64.64 & 31.53 & 11.49 & \secondbest{69.02} & 33.84 & 13.37 \\ 
        \midrule
        \band \ourmethod \emph{(ours)} & \secondbest{56.92} & \best{67.58} & \best{39.20} & \best{13.91} & \best{71.41} & \secondbest{41.33} & \best{15.88} \\ 
        \bottomrule
\end{tabularx}
\vspace{-0.5em}
\end{table}

\subsection{A joint evaluation  protocol}\label{sec:exp:awga}
\begin{figure}[t]
    \centering
    \includegraphics[width=\linewidth]{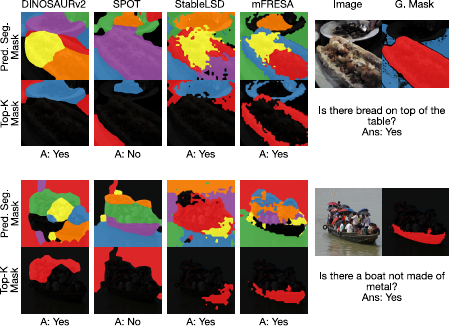}
       \caption{\textbf{Qualitative examples.} AwGA scores each example using the image, grounding masks (G. Mask), and the question–answer pair. The need for AwGA is evident: DINOSAURv2 achieves high G-Acc (correct ans. and high mIoU) but low AwGA, since the Top-$K$ answer-attributed slots poorly overlap with the grounded mask. Also, StableLSD predicts the correct answer, but has a low AwGA due to weak grounding overlap.}
       \vspace{-0.75em}
    \label{fig:exp:gqa}
\end{figure}

To assess OCL models using our proposed AwGA metric, we use the GQA validation set \citep{hudson2019gqa}, a large-scale VQA dataset with grounding boxes for each question. To better align with our evaluation, we enhance GQA by converting bounding-box annotations into masks using SAM2 \citep{ravi2024sam2}, treating boxes as prompts. To ensure that grounded objects are salient, we filter out images with more than seven boxes or those covering less than 10\% of the image area. We use the same mask-generation pipeline for all models to ensure a fair comparison. We call this dataset the enhanced GQA dataset (eGQA). Some examples are shown in \cref{fig:exp:gqa} and more details are provided in \cref{sec:appendix:eGQA}.

We report accuracy, mIoU, G-Acc, and our proposed AwGA metric in \cref{tab:exp:awga_metric}. mIoU measures the overlap between predicted and ground-truth masks, but by itself, cannot determine whether the correct answer is grounded in the appropriate object slots. G-Acc penalizes poor localization but overlooks fragmented or incorrect slot usage (\tptwo). By contrast, AwGA jointly evaluates both localization and representation usefulness under a single VLM probing protocol, providing a grounded measure that penalizes both localization and representation fragmentation (see \cref{fig:exp:metrics_visual}). 
Interestingly, models with top object discovery or accuracy scores are not always SOTA under G-Acc or AwGA, underscoring the pitfalls of disjoint evaluation. 

\begin{takeaway}
    \emph{Takeaway 4.} Using only object discovery or semantic prediction performance metrics provides an incomplete evaluation of OCL models. AwGA complements these metrics by jointly measuring \emph{what} and \emph{where}, and penalizes both localization and representation fragmentation.
\end{takeaway}

\subsection{Ablations}
\begin{figure}[ht]
\centering
\begin{tabular}{cc}
     \includegraphics[height=0.3\linewidth]{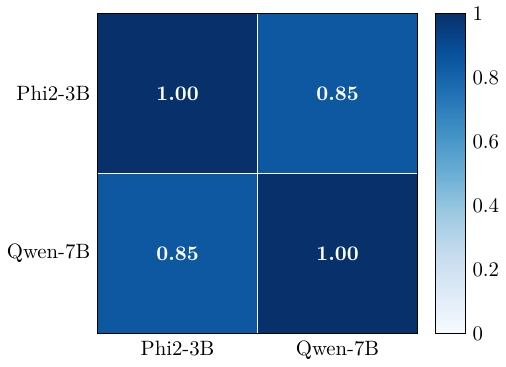} &  
     \includegraphics[height=0.3\linewidth]{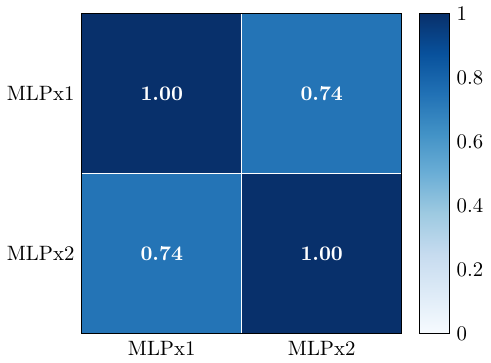}
\end{tabular}
\caption{\textbf{Robustness of AwGA.} Spearman's rank correlation for the AwGA metrics for different LLM and connectors designs. AwGA remains stable across different LLMs and connector architectures, suggesting that our VLM-based evaluation is relatively insensitive to specific LLMs or connector architectures.}
\label{fig:exp:robust_awga}
\vspace{-0.75em}
\end{figure}

\myparagraph{Effect of LLM and connector.}
We next show that the AwGA metric is robust to both LLM and connector choices. To evaluate this, we compute the Spearman rank correlation of AwGA scores between Phi-2 (3B) and Qwen2 (7B) on the enhanced grounded GQA dataset. Then, using Phi-2 as the LLM, we evaluate against two popular connector variants, 1-layer MLP (MLP$\times$1) and 2-layer MLP (MLP$\times$2), and again report Spearman correlations. As seen in \cref{fig:exp:robust_awga}, the rank correlations remain consistently strong (>0.70), showing that AwGA rankings are stable across small and large-sized LLMs and different connectors, indicating the robustness of our evaluations. Additional robustness results are provided in \cref{tab:appx:corr}.

\myparagraph{Robustness of AwGA to the attribution method.}
The AwGA metric requires selecting an attribution method as the \emph{only} hyperparameter. In this experiment, we compare two common attribution methods: simply using the gradient information associated with each slot, and integrated gradients \citep{sundararajan2017axiomatic}. As seen in \cref{tab:abl2}, the AwGA metric is very robust to the choice of attribution method (Spearman rank correlation of 0.77). We use gradient-based attribution for computational efficiency.

\begin{table}[ht]
\centering
\caption{\textbf{Attribution robustness.} AwGA is robust to attribution choice, showing a strong Spearman correlation ($\rho=0.77$) between gradient and integrated-gradient attributions.}
\label{tab:abl2}
\scriptsize
\vspace{-0.5em}
\begin{tabularx}{\linewidth}{@{}X*{2}{S[table-format=2.2,table-column-width=1.2cm]}@{}}
        \toprule
        & \multicolumn{2}{c}{\textbf{Attr. Type}} \\\cmidrule(lr){2-3}
         & {\textbf{Grad.}} & {\textbf{Int. Grad.}} \\
        \midrule
        DINOSAUR & 11.64 & 7.58 \\
        DINOSAURv2 & 11.92 & 7.87 \\
        FT-DINOSAUR &  \U {12.81} &  8.26 \\
        \midrule
        SlotDiffusion & 12.42 & \U {8.83} \\
        StableLSD  & 11.49 & 8.04 \\
        \midrule
        \band \ourmethod (ours) & \B 13.91 & \B 9.48 \\
        \bottomrule
    \end{tabularx}

\end{table}

\myparagraph{Choice of reconstruction objective.}
\begin{table}[t]
\centering
\scriptsize
\setlength{\tabcolsep}{1pt}
\begin{minipage}[t]{1\linewidth}
\captionsetup{type=table} 
\captionof{table}{\textbf{Importance of HOG and feature decoders.}
Both HOG and (DINOv2) feature decoders improve localization performance (mIoU), downstream VQA performance, and joint evaluation metric, indicating their importance. All experiments in (\%, $\uparrow$).}
\label{tab:appx:mFresa_abl}
\vspace{-0.5em}
\begin{tabularx}{\textwidth}{@{}ccc*{4}{>{\centering\arraybackslash}X}|*{2}{>{\centering\arraybackslash}X}@{}}
\toprule
\multicolumn{3}{c}{\textbf{Decoder}}  &  \textbf{GQA} & \textbf{OOD} & \textbf{Sugar C.}& \textbf{VQAv2} & \textbf{mIoU}& \textbf{AwGA} \\ 
Img. & Feat. & HOG & & & & & & \\
\midrule
\cmark & -- & -- & 51.45	& 52.89	& 72.92  &	\secondbest{62.06} & 47.94  & 11.49 \\

\cmark & \cmark & -- & \secondbest{52.21}	& \secondbest{54.20}	& \secondbest{75.92}	& 61.40 &  \secondbest{55.76}  & \secondbest{13.37} \\
\band \cmark & \cmark & \cmark & \best{53.90}	& \best{55.27}	& \best{77.27}&	\best{65.58} & \best{56.92}  & \best{13.91}  \\
\bottomrule
\end{tabularx}
\end{minipage}
\vspace{-0.75em}
\end{table}

\ourmethod builds on StableLSD and includes two new decoders: the feature and HOG decoders. In \cref{tab:appx:mFresa_abl}, we quantify the effect of each decoder. We observe that simply adding the DINOv2 feature decoder improves results across almost all tasks (localization and representation usefulness). Additionally, incorporating HOG features further enhances performance for both abilities, underscoring the effectiveness of both decoders in learning more accurate slot representations that provide better localization and more useful slot representations. 

\myparagraph{Limitations.}
While our evaluation framework and AwGA provide a broader view of object-centric learning (OCL), they have some limitations. First, our protocol relies on training large vision-language models (VLMs) as evaluators; although this cost is amortized across many benchmarks, the one-time training for each OCL model is still expensive. Second, AwGA requires access to grounding masks for question--answer pairs, which currently limits its evaluation to our eGQA benchmark. Nevertheless, we view AwGA as an important additional evaluation protocol for evaluating OCL models. Finally, our work focuses on images, leaving extensions to video-based OCL models \citep{elsayed2022savi,kipf2022conditional} as future work.

\section{Conclusion}
Object-centric learning (OCL) has made notable progress in unsupervised object discovery (UOD) for real-world scenes. However, broader goals such as compositionality, counterfactual reasoning, and OOD robustness remain underexplored, partly because existing evaluation schemes require costly task-specific retraining.
To address this, we propose a scalable evaluation protocol based on visual instruction tuning of an LLM. Our protocol evaluates how effectively a VLM can leverage slot representations from different OCL models across diverse VQA benchmarks, without training task-specific probes for each benchmark. We find that current OCL models, though competitive, still lag the DINOv2 encoder in absolute performance on several key benchmarks. This indicates the need for developing stronger OCL models. We then show the need for an evaluation metric that jointly evaluates the localization and representation usefulness of OCL models using a single unified metric. To this end, we introduce a grounded evaluation benchmark (eGQA) and propose Attribution-aware Grounded Accuracy (AwGA), a unified metric that jointly evaluates the ``what'' and ``where'' aspects of OCL. Finally, we include a simple multi-target reconstruction baseline (\ourmethod) as a reference, demonstrating that combining reconstruction objectives improves reasoning and localization capabilities.

\section*{Impact Statement}
\myparagraph{Positive.} Object-centric learning (OCL) aims to represent scenes as sets of objects, which may support more structured and interpretable perception \citep{baillargeon1985object,teglas2011pure}. This work contributes an evaluation framework that benchmarks the downstream utility of OCL representations across diverse reasoning-oriented VQA benchmarks (\eg, compositionality, OOD generalization, and robustness) without training task-specific probes. We further identify fragmentation issues in existing evaluation protocols and propose eGQA, together with the AwGA metric, to jointly evaluate the \emph{what} and \emph{where} capabilities of OCL models. We hope these tools will support more reliable evaluation and accelerate progress toward OCL models that improve both the localization and the usefulness of learned object representations.

\myparagraph{Negative.} Our evaluation methodology for the OCL models is based on VLMs, which employ a pre-trained large language model (LLM). Thus, our evaluation can also inherit LLM biases; therefore, evaluating vision encoders with LLMs should be done cautiously. These biases can be eliminated by using and training VLM on a more balanced dataset. However, this is out of the scope of this work. Another concern with our evaluation framework is that training the VLM for evaluating the OCL models is not energy efficient, as standard metrics do not require such training. However, it should be noted that the amortized cost of evaluating OCL models within our evaluation framework is still lower than that of training separate linear or transformer-based probes to assess the capabilities of OCL methods.

\bibliography{main}
\bibliographystyle{icml2026}

\clearpage
\newpage
\appendix
\onecolumn
\section{\ourmethod: An improved baseline}
We here provide additional details of our proposed baseline, \ourmethod, which builds upon the StableLSD framework \citep{jiang2023object}. StableLSD is an encoder-decoder architecture with a slot-attention bottleneck. It employs a DINOv2 model as the encoder, and a frozen Stable Diffusion  \citep{rombach2022high} model as the decoder. The slot attention module is trained using an image reconstruction loss. 
We extend this design by introducing two additional decoders: a HOG feature decoder and a DINOv2 feature decoder.  

Given the slots, the \emph{HOG decoder} reconstructs the HOG feature map of the input image \citep{dalal2005histograms}, encouraging slots to better capture object boundaries via edge information. HOG features are computed by aggregating gradient orientations within local neighborhoods. The \emph{DINOv2 feature decoder}, inspired by DINOSAUR \citep{seitzer2022bridging}, reconstructs DINOv2 features from the slots, complementing image-level supervision.  
The overall training objective is given as
\begin{equation}
    \mathcal{L} = L_2({I}_\text{inp}, {I}_\text{recon}) +
    L_2({F}_\text{inp}, {F}_\text{recon}) +
    L_2({H}_\text{inp}, {H}_\text{recon}),
    \label{eq:meth:loss}
\end{equation}
where ${I}$ denotes images, ${F}$ DINOv2 features, and ${H}$ HOG features of the input and reconstruction, respectively.  

\begin{figure}[ht]
    \centering
    \includegraphics[width=0.90\linewidth]{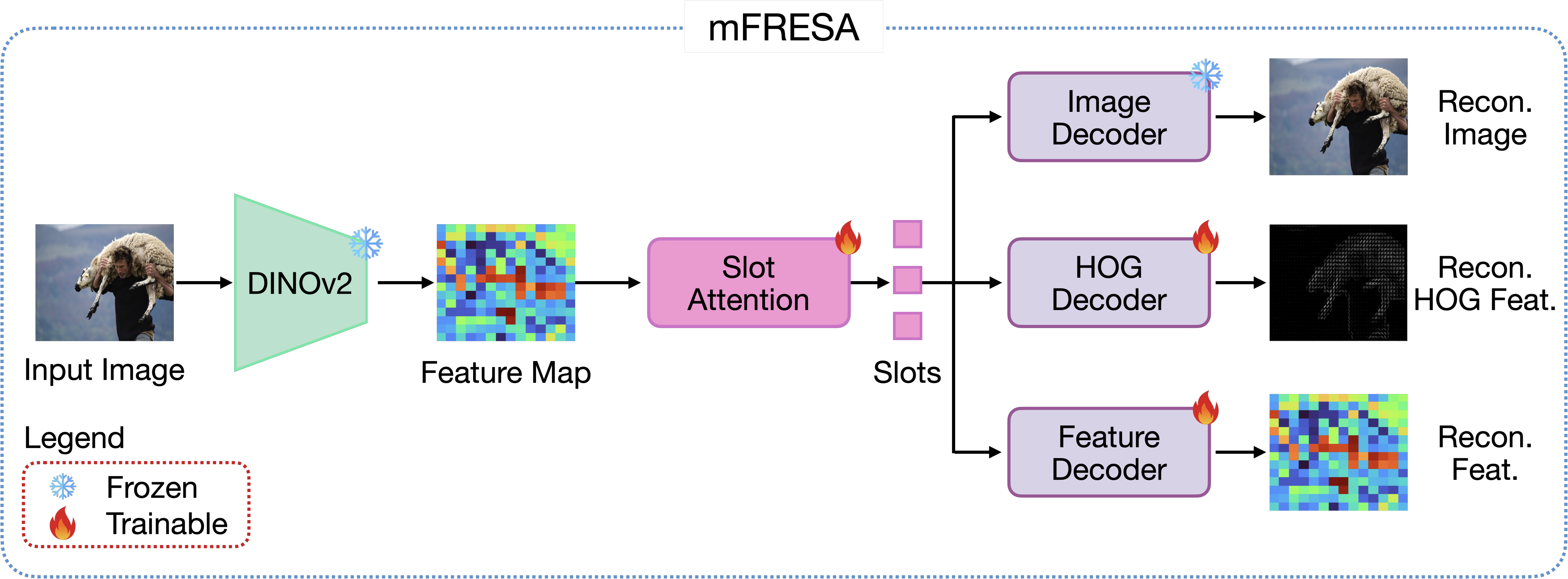}
    \caption{\textbf{Multi-feature reconstruction for slot attention (\ourmethod)} uses a DINOv2 \citep{oquab2023dinov2} model as a feature encoder network. The slot-attention module groups the obtained features into slots. Multiple decoders reconstruct the image,  HOG features, and DINOv2 features from the slots. The slot-attention module and the HOG and feature decoders are trainable, while DINOv2 and the image decoder model (a diffusion decoder) are kept frozen following StableLSD. The model is trained with \cref{eq:meth:loss}. Not visualised: The HOG features are computed as described in \citep{dalal2005histograms}.
    }
    \label{fig:method:network_main}
\end{figure}

The key contribution of \ourmethod is the \emph{joint reconstruction of image, feature, and edge signals}, enabling slots to learn stronger object-centric representations. A detailed network diagram is shown in \cref{fig:method:network_main}.

\label{sec:appx:method_train_detail}
\newcolumntype{Y}{>{\centering\arraybackslash}X}
\newcolumntype{L}{>{\raggedright\arraybackslash}X}
\begin{table}[t]
\centering
\scriptsize
\caption{\textbf{Architectural and training details for \ourmethod.}}
\begin{tabularx}{\textwidth}{@{}lLc@{}}
\toprule
\textbf{Module}            &  \textbf{Hyperparameter}             &  \textbf{Value} \\ 
\midrule
\multirow{11}{*}{General}     &   Batch size       & 32  \\     
            &   Precision        & fp16    \\
            &   Learning rate    & 2e-5      \\
            & Learning rate scheduler & Constant \\
            &   Optimizer        & Adam \citep{kingma2014adam}    \\
            & Adam ($\beta_1$, $\beta_2$) & (0.9, 0.999) \\
            & Adam eps & 1e-8 \\
            & Weight decay & 1e-2 \\
            &   Learning rate scheduler & Constant\\
            &   Iterations & 500K \\
            & Max. grad norm & 1.00 \\
\midrule
\multirow{4}{*}{Encoder}     &   Architecture             & DINOv2 \citep{oquab2023dinov2}  \\
            &   Patch size       & 14       \\
            &   Backbone         & ViT-B \citep{dosovitskiy2020image}    \\
            &   Embedding dimensions   & 768   \\
\midrule
\multirow{3}{*}{Slot Attention} & \# Iterations & 5 \\
               & \# Slots & 7 \\
               & Slot Size & 768 \\
\midrule
\multirow{2}{*}{Image Decoder}     & Architecture               & Stable Diffusion \citep{rombach2022high}\\
            & Model version         & 2.1         \\
\midrule
            \multirow{2}{*}{Feat. Decoder} & Architecture & MLP \\
            & No.\ of layers & 2  \\
            & Hidden dimensions & 1536 \\
\midrule
            \multirow{2}{*}{HOG Decoder} & Architecture & MLP \\
            & No.\ of layers & 2  \\
            & Hidden dimensions & 1536 \\
\bottomrule

\end{tabularx}

\label{tab:appendix:training_detail}
\end{table}

\ourmethod is trained on a single NVIDIA A100 GPU with 80GB of VRAM. The encoder and image decoder components closely follow the StableLSD setup \citep{jiang2023object}, with \ourmethod introducing two additional modules: a HOG feature extractor \citep{dalal2005histograms,wei2022masked} and decoder, as well as a DINOv2 feature decoder. The model is trained for 500K iterations on the COCO dataset \citep{lin2014microsoft}. Images fed to the DINOv2 encoder \citep{oquab2023dinov2} are resized and center-cropped to 518$\times$518 pixels. We used an Adam optimizer \citep{kingma2014adam} for training our model. The full training and architectural details of our method are shown in \cref{tab:appendix:training_detail}. The additional decoders also lead to additional training times, for example, StableLSD on a single A100 GPU trains in approx. 43 hours. Adding the feature decoder increases the training time to 65 hours. Adding the HOG decoder further increases the total training time to 82 hours.

\section{Additional results}
\subsection{Additional application of our evaluation framework}
\textbf{Quantifying the type of learned slots.} 
Our evaluation framework can be used to 
to probe whether architectural choices bias slots toward encoding specific properties (\eg, spatial, relational, or object). The GQA dataset \citep{hudson2019gqa} categorises questions into four semantic types: \emph{(1)} object (existence), \emph{(2)} attribute (properties or position), \emph{(3)} category (class membership), and \emph{(4)} relation (subject--object relations). 
As shown in \cref{tab:exp:type_slot_abl}, feature reconstruction methods excel at existence and relation questions, whereas image-only methods like Slot Diffusion \citep{wu2023slotdiffusion} and StableLSD \citep{jiang2023object} lag behind. \ourmethod, which combines both, achieves the best results in three of four categories. For MM-Vet \citep{yu2023mm}, covering recognition and spatial queries, results are mixed: feature- and image-based approaches perform similarly on recognition, while Slot Diffusion performs best on spatial relations.
\begin{table}[t]
\centering
\scriptsize
\setlength{\tabcolsep}{1pt}
\begin{minipage}[t]{1\textwidth}
\captionsetup{type=table} 
\captionof{table}{
\textbf{Type of properties encoded by slots}. VLM-based evaluation (accuracy in \%, $\uparrow$) allows us to quantify the type of properties that a slot encodes via a categorization of the questions.}
\label{tab:exp:type_slot_abl}
\begin{tabularx}{\textwidth}{@{}l*{4}{>{\centering\arraybackslash}X} *{2}{>{\centering\arraybackslash}X}@{}}
\toprule
\textbf{Dataset} & \multicolumn{4}{c}{\textbf{GQA} \citep{hudson2019gqa}} & \multicolumn{2}{c}{\textbf{MM-Vet} \citep{yu2023mm}} \\
\cmidrule(lr){2-5} \cmidrule(lr){6-7}
 & Attribute & Category & Object & Relation & Recognition & Spatial \\
\midrule
DINOSAURv2 \citep{seitzer2022bridging} & \secondbest{57.58} & \secondbest{45.26} & \secondbest{78.02} & \secondbest{46.95} & 21.5 & \secondbest{23.9} \\
FT DINOSAUR \citep{didolkar2025zero} & 56.77 & 43.17 & \best{79.95} & 45.54 & 18.7 & 19.9 \\
SPOT \citep{kakogeorgiou2024spot} & 57.15 & 42.47 & 75.06 & 43.24 & 19.7 & 23.1 \\
\midrule
Slot Diffusion \citep{wu2023slotdiffusion} & 57.40 & 39.77 & 73.52 & 41.28 & 20.5 & \best{28.9} \\
StableLSD \citep{jiang2023object} & 56.73 & 43.43 & 75.19 & 44.20 & \secondbest{21.6} & 22.3 \\
\band \ourmethod \emph{(ours)} & \best{59.08} & \best{46.74} & 77.63 & \best{47.23} & \best{21.9} & 22.1 \\
\bottomrule
\end{tabularx}
\end{minipage}
\end{table}

\myparagraph{Correlation analysis.}
We evaluate the robustness of our slot-attention evaluation framework to the choice of the large language model (LLM). 
While the choice of LLMs affects the absolute performance of the models, we find that the relative ranking of OCL models remains largely unchanged across different LLMs. 
\Cref{tab:appx:corr} reports Spearman’s rank correlations between results obtained with Phi2 and Qwen2-7B across multiple datasets. 
The strong correlations ($\rho \geq 0.89$) indicate that our proposed VLM-based evaluation framework is stable, with model rankings preserved across LLMs.

\begin{figure}[b]
    \centering
    \includegraphics[width=\linewidth]{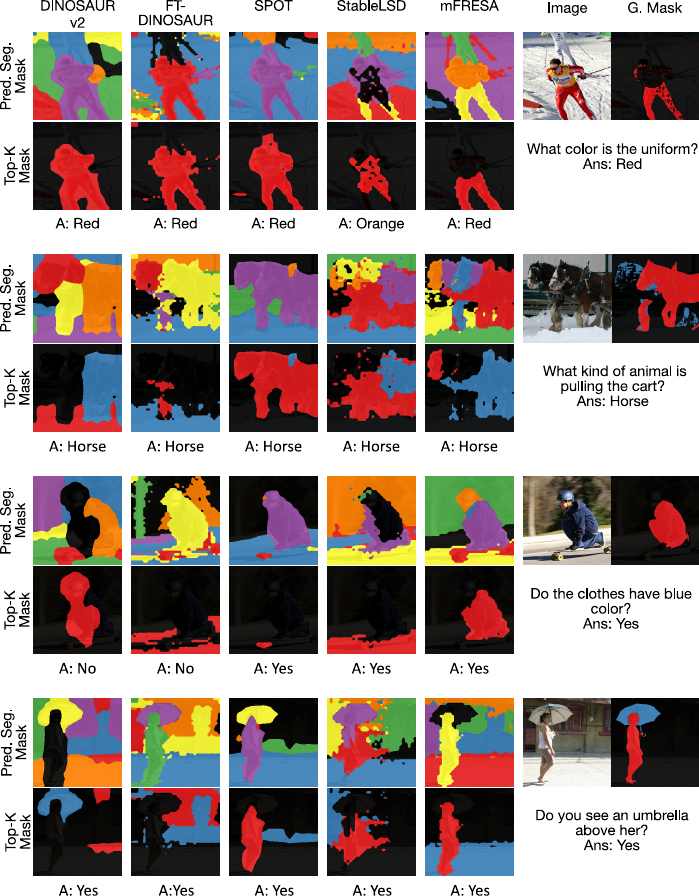}
    \caption{\textbf{Qualitative examples.} We visualize predicted masks and attribution-aware masks obtained by selecting the $K$ tokens with the highest attributions. A good predicted mask does not necessarily imply a correct slot encoding, which can lead to wrong answers (See text). G. Mask denotes the ground-truth eGQA mask for the regions required to answer each question. Pred. Seg. Mask denotes the predicted mask from the slot attention module.}
    \label{fig:exp:qual_samples}
    \vspace{-0.75em}
\end{figure}

\begin{table}[h]
\centering
\caption{\textbf{Spearman rank correlations between different models when using Phi2 and Qwen2-7B models as LLMs.} The results show that our evaluation framework is robust to the choice of LLM, and the rank between the models remains largely preserved (very strong correlation). Spearman $\rho$ (max 1, $\uparrow$).}
\label{tab:appx:corr}
\scriptsize
\begin{tabularx}{\linewidth}{@{}l
    *{11}K
    @{}}
\toprule
 \textbf{Dataset}& \textbf{GQA} & \textbf{POPE} & \textbf{MME} & \textbf{MMVet} & \textbf{VQAv2} & \textbf{OOD} & \textbf{Nat. Bench} & \textbf{Sugar C.} & \textbf{AwGA} \\
\midrule
Spearman $\rho$ & 0.98 & 0.95 & 0.70 & 0.76 & 0.98 & 0.86 & 0.91 & 0.85 & 0.92 \\
\bottomrule
\end{tabularx}
\end{table}
\myparagraph{Qualitative results.} \cref{fig:exp:qual_samples} presents qualitative examples comparing the predicted masks with attribution-aware masks obtained by selecting the slots with the $K$-highest attributions based on a gradient-based attribution method \citep{simonyan2013deep}. These results highlight that high-quality predicted masks or accuracy alone do not always imply the usefulness of the slot representations. 
For example, in \emph{top example}, StableLSD, despite having a good predicted mask, produces an incorrect output. Importantly, in \emph{example-3}, SPOT and StableLSD produce a correct response to the question, but use the wrong slot (as seen by the Top-k mask) to answer the question.

\myparagraph{Mean and standard deviation of results.}
Training the VLMs with different random seeds and evaluating the resulting models is computationally intensive, as these VLMs are trained across 8 NVIDIA A100 GPUs, with the fine-tuning stage typically taking around 24 hours. This makes it infeasible to provide results from multiple runs using this approach.
Instead, we report mean and standard deviation results for \ourmethod and several baseline models on representative datasets during \emph{evaluation}. We set the temperature for LLM generation to 0.02 and averaged the results across five random seeds (42, 1337, 2025, 4378, 8921). We report the results on SugarCrepe \citep{hsieh2023sugarcrepe}, MME \citep{fu2024mme}, and POPE \citep{li2023evaluating} as representative datasets for visual question answering in \cref{tab:appendix:error_bars}. 
We use the accuracy as an evaluation metric for the SugarCrepe and POPE datasets. For the MME dataset, we provide scores based on the MME evaluation script (with 2000 as the maximum for the perception task).
Please note that the numbers reported in \cref{tab:exp:main_cmp_vqa} and \cref{tab:exp:robust_vqa} of the main paper are for a temperature value set to 0. Comparing these to \cref{tab:appendix:error_bars}, we observe the ranking of the models following the same trend as with temperature 0. Setting the temperature $> 0$ introduces randomness into the output of large language models, enabling us to obtain the mean and standard deviation during evaluation.
We observe small performance variation and consistent model rankings across random seeds, indicating the robustness of our VLM-based evaluation framework.

\begin{table}[h!]
\centering
\scriptsize
\caption{\textbf{Mean and standard deviation of results.} Performance comparison with mean and standard deviation of different methods in a selection of representative datasets. SugarCrepe \citep{hsieh2023sugarcrepe} and POPE \citep{li2023evaluating} are evaluated in terms of accuracy (in \%, $\uparrow$), MME \citep{fu2024mme} in terms of its score ($\uparrow$). We observe small performance variation and consistent model rankings across random seeds, indicating the robustness of our VLM-based evaluation framework.
We highlight the
\best{best} and \secondbest{second best} model among slot-attention methods.}
\label{tab:appendix:error_bars}
\smallskip
\begin{tabularx}{\textwidth}{@{}l*{3}{Y}@{}}
\toprule
\textbf{Method} &  \textbf{SugarCrepe} \citep{hsieh2023sugarcrepe} &  \textbf{MME} \citep{fu2024mme} &  \textbf{POPE} \citep{li2023evaluating} \\
\midrule
DINOv2 & 82.14 $\pm$ 0.13 &  1283.96 $\pm$ 07.29 &  82.08 $\pm$ 0.10 \\
\midrule
DINOSAURv2 &  \secondbest{76.20 $\pm$ 0.25} & 1123.47 $\pm$ 12.63 & \secondbest{81.84 $\pm$ 0.24} \\
FT-DINOSAUR & 71.25 $\pm$ 0.23 & 1016.15 $\pm$ 22.11 & 81.54 $\pm$ 0.18 \\
SPOT & 71.65 $\pm$ 0.15 &  1066.04 $\pm$ 07.19 & 79.69 $\pm$ 0.04 \\
\midrule
Slot Diffusion & 70.23 $\pm$ 0.33 &  1090.75 $\pm$ 08.09 & 79.74 $\pm$ 0.12 \\
StableLSD & 72.89 $\pm$ 0.32 & \secondbest{1126.06 $\pm$ 17.21} & 81.13 $\pm$ 0.11 \\
\band \ourmethod \emph{(ours)} & \best{77.18 $\pm$ 0.27} &  \best{1184.40 $\pm$ 19.52} &  \best{82.20 $\pm$ 0.08} \\
          
\bottomrule
\end{tabularx}

\end{table}

\section{Datasets}
\label{appendix:sec:dataset}
Here we describe the datasets used in \cref{sec:exp:eval_standard} for our VQA-based evaluation of OCL models. %

\myparagraph{VQAv2.0} \citep{goyal2017making} is a dataset of 265,016 images from COCO and abstract scenes, each paired with an average of 5.4 open-ended questions requiring vision, language, and commonsense reasoning. Each question includes 10 ground-truth answers and 3 plausible but likely incorrect answers, making it a robust benchmark for evaluating visual question-answering (VQA) models.
 
\myparagraph{GQA} \citep{hudson2019gqa} is a VQA dataset for real-world images that requires visual, spatial, and compositional reasoning. Importantly, GQA provides grounding masks (referred objects for answering questions) for each question in the validation set.

\myparagraph{POPE} \citep{li2023evaluating}. The Polling-based Object Probing Evaluation (POPE) assesses object-level perception and hallucination in vision-language models by querying whether specific objects are present in images. It consists of three settings: \emph{(i)} Random -- this setting samples absent objects at random, \emph{(ii)} Popular -- this setting selects missing objects from a frequently occurring object pool, and \emph{(iii)} Adversarial -- this setting targets commonly co-occurring but visually absent objects to challenge the model’s grounding ability. In total, POPE consists of 3 sets of image-question pairs, each containing 1500 pairs with answer ``Yes'' and 1500 pairs with answer ``No''.

\myparagraph{MME} \citep{fu2024mme} is a comprehensive benchmark designed to evaluate the capabilities of multimodal large language models (MLLMs) across 14 diverse subtasks spanning both perception and cognition. In our work, we focus specifically on perception tasks, including coarse-grained recognition (existence, count, position, colour), fine-grained recognition (poster, celebrity, scene, landmark, artwork), and optical character recognition (OCR). Model performance on these tasks is measured using the perception score, capped at 2000 points.

\myparagraph{MM-Vet} \citep{yu2023mm}. Unlike standard evaluation benchmarks, MM-Vet evaluates the integration of key vision-language (VL) capabilities, such as recognition, optical character recognition (OCR), knowledge reasoning, language generation, spatial understanding, and mathematical reasoning. MM-Vet contains 200 images and 218 questions, each paired with its respective ground truth.

\section{Enhanced GQA dataset}
\label{sec:appendix:eGQA}
We construct our enhanced GQA (eGQA) dataset used in \cref{sec:exp:awga} based on the validation split of the original GQA dataset \citep{hudson2019gqa}. Our enhanced version comprises 10,000 questions, each accompanied by grounded segmentation masks. To convert grounding bounding boxes—\ie, the coordinates of objects referenced in the questions—into segmentation masks, we utilise the SAM2 model \citep{ravi2024sam2}, specifically the ``sam2.1-heira-large'' checkpoint with its default configuration.

\begin{figure}
    \centering
    \includegraphics[width=1\textwidth]{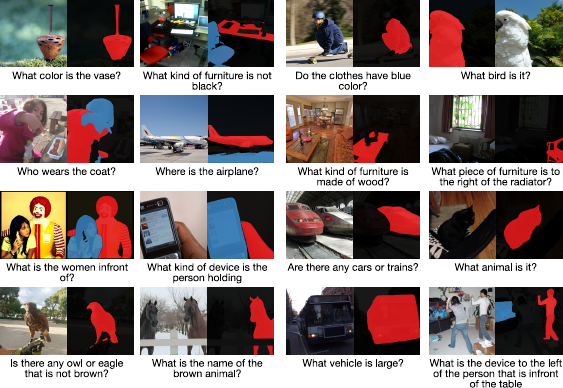}
    \caption{\textbf{Additional samples from the enhanced GQA dataset.} The dataset is composed of questions and answers (not shown) pairs. The grounding masks denote the masks of the referring objects required to answer the question.}
    \label{fig:appendix:egqa_samples}
    
\end{figure}

To ensure relevance and clarity, we apply filtering criteria that discard images with more than 7 bounding boxes or with total box coverage less than 10\% of the image area. These thresholds are chosen to retain only prominent objects while maintaining compatibility with object-centric learning (OCL) models trained on the COCO dataset \citep{lin2014microsoft}, which typically utilize seven slots. Additional examples from our eGQA dataset are shown in 
\cref{fig:appendix:egqa_samples}.

\end{document}